\pdfoutput=1

\documentclass[11pt]{article}

\usepackage[final]{acl}

\usepackage{times}
\usepackage{latexsym}

\usepackage[T1]{fontenc}

\usepackage[utf8]{inputenc}

\usepackage{microtype}

\usepackage{inconsolata}

\usepackage{graphicx}

\usepackage{amsmath}
\usepackage{mathrsfs}
\usepackage{booktabs}    
\usepackage{amssymb}
\usepackage{amsthm}
\usepackage{subcaption}
\usepackage{mathtools}
\usepackage{multirow}
\usepackage{float}

\usepackage{algorithmic}
\usepackage{algorithm}

\title{TReMu: Towards Neuro-Symbolic Temporal Reasoning for LLM-Agents with Memory in Multi-Session Dialogues}

\author{Yubin Ge\textsuperscript{$1,2$}\thanks{Work done as an intern at AWS}, Salvatore Romeo\textsuperscript{$2$}, Jason Cai\textsuperscript{$2$}, Raphael Shu\textsuperscript{$2$}, Yassine Benajiba\textsuperscript{$2$}\thanks{Work done while at AWS, now at Oracle}, \\ \textbf{Monica Sunkara\textsuperscript{$2$}}, \textbf{Yi Zhang\textsuperscript{$2$}} \\
\textsuperscript{$1$}University of Illinois Urbana Champaign, USA \\
\textsuperscript{$2$}Amazon Web Services \\
\texttt{\{yubinge, romeosr, cjinglun, zhongzhu, sunkaral, yizhngn\}@amazon.com}}

\begin{document}
\maketitle
\begin{abstract}
     Temporal reasoning in multi-session dialogues presents a significant challenge which has been under-studied in previous temporal reasoning benchmarks. To bridge this gap, we propose a new evaluation task for temporal reasoning in multi-session dialogues and introduce an approach to construct a new benchmark by augmenting dialogues from LoCoMo and creating multi-choice QAs. Furthermore, we present TReMu, a new framework aimed at enhancing the temporal reasoning capabilities of LLM-agents in this context. Specifically, the framework employs \textit{time-aware memorization} through timeline summarization, generating retrievable memory by summarizing events in each dialogue session with their inferred dates. Additionally, we integrate \textit{neuro-symbolic temporal reasoning}, where LLMs generate Python code to perform temporal calculations and select answers. 
    Experimental evaluations on popular LLMs demonstrate that our benchmark is challenging, and the proposed framework significantly improves temporal reasoning performance compared to baseline methods, raising from 29.83 on GPT-4o via standard prompting to 77.67 via our approach and highlighting its effectiveness in addressing temporal reasoning in multi-session dialogues.N
\end{abstract}
\section{Introduction}

Temporal reasoning, a fundamental cognitive ability, plays a crucial role in human perception \cite{hoerl2019thinking}. 
It encompasses not only basic concepts such as ordering and duration but also more complex functions like task planning and causal relation discovery \cite{vila1994survey}. Besides, large language models (LLMs) have brought significant advancements to various tasks. These models typically undergo pre-training on extensive text corpora, endowing them with unprecedented accuracy in predicting the next token given some context \cite{ge2023supervised}.
Despite the remarkable advancements in LLMs and their emerging reasoning capabilities \cite{huang2023towards}, previous studies have shown that their performance in temporal reasoning remains suboptimal \cite{chu-etal-2024-timebench,qiu2024large}.

In the context of multi-session dialogues, temporal reasoning is both critical and challenging for LLM-agents. As dialogue sessions proceed, storing and retrieving relevant information efficiently becomes more difficult \cite{maharana-etal-2024-evaluating},
such as failing to retrieve specific temporal details from long history and dialogues exceed the input limit of LLMs. Additionally, research has shown that LLMs overlook important contextual information from long dialogue histories due to the accumulation of irrelevant historical data, referred to as "historical noise" \cite{wang2023enhancing}. 
These challenges underscore the need for enhanced temporal reasoning capabilities in LLM-agents for effective handling of multi-session dialogues.


\begin{figure}  
\centering
\includegraphics[width=\linewidth]{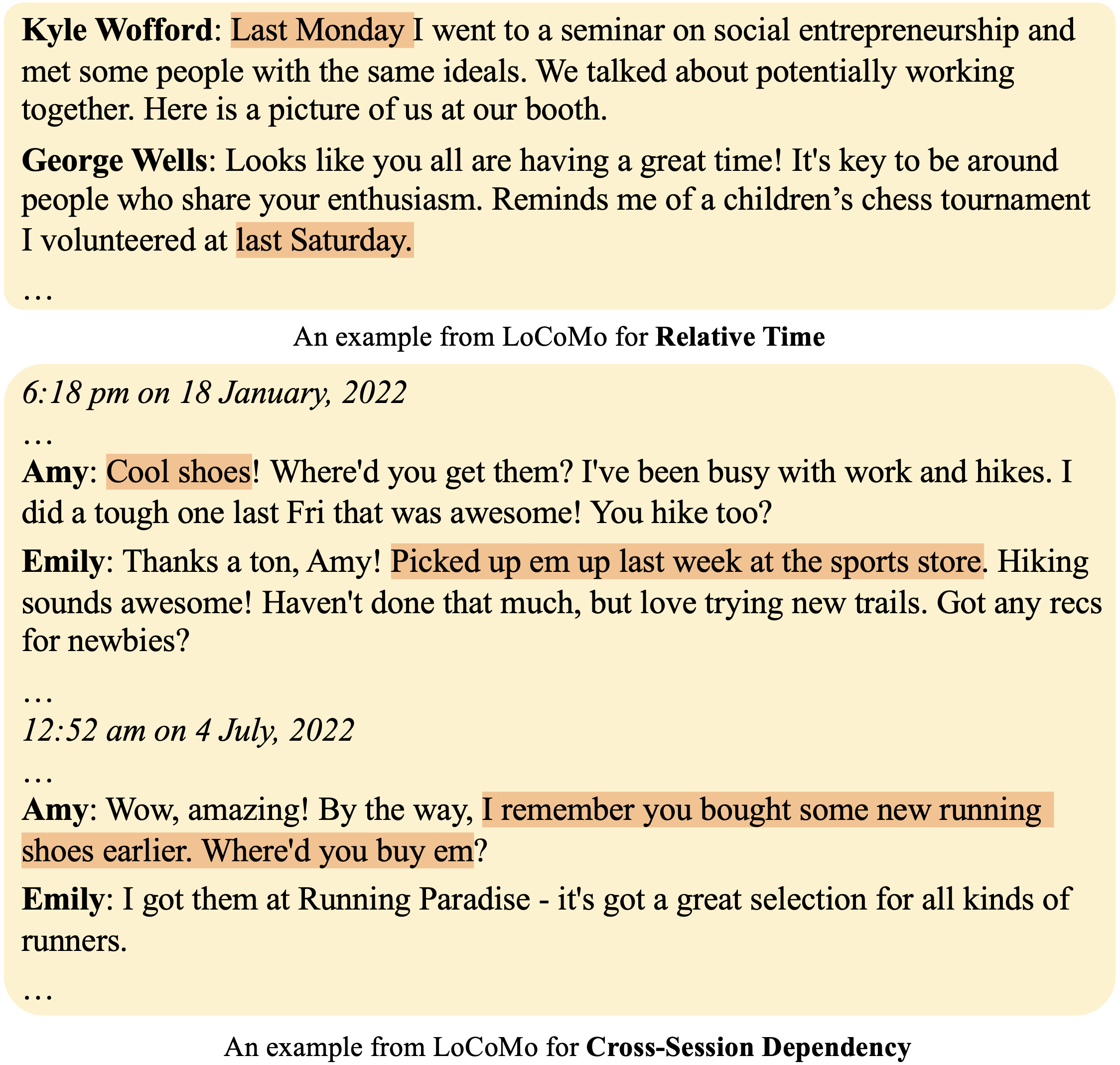}
    \caption{Examples from LoCoMo showing the two temporal characteristics we focus on in this work.} 
    \label{fig:example}
\end{figure}

However, most existing temporal reasoning benchmarks cannot be used directly for this study, because they are usually built on shorter texts, such as stories and Wikipedia articles, that contain clear temporal information \cite{chen2dataset,wang-zhao-2024-tram,xiong-etal-2024-large}. Even benchmarks designed for dialogues, like TimeDial \cite{qin2021timedial} and LoCoMo \cite{maharana-etal-2024-evaluating}, do not explicitly consider the special temporal characteristics in multi-session dialogues, such as \textit{relative time} and \textit{cross-session dependency}. For instance, speakers often use relative time expressions instead of specific dates, requiring the model to infer exact event times. Moreover, it is common for speakers to recall past events from previous sessions, creating cross-session dependencies, where events from different sessions involve the same or related entities and reflect changes over time. This further requires LLMs to retain context effectively when reasoning about events across multiple sessions.

In this work, we present \textbf{TReMu} (\textbf{T}emporal \textbf{Re}asoning for LLM-Agents in \textbf{Mu}lti-Session Dialogues), a novel framework designed to enhance temporal reasoning in multi-session dialogues. Our framework introduces \textbf{time-aware memorization}, which uses timeline summarization to generate summaries for each dialogue session, identifying events and associating them with their inferred dates. These summaries, linked to specific times (either session times or inferred event dates), serve as retrievable memory. This effectively addresses events expressed in relative time by distinguishing when such an event occurred from when it was mentioned by the speaker.

During reasoning, we propose a \textbf{neuro-symbolic temporal reasoning} approach inspired by recent work that integrates LLMs with symbolic reasoning, translating questions into symbolic language before using a solver to find answers \cite{pan2023logic,olausson2023linc}. Specifically, given a temporal question, we retrieve relevant memory and instruct the LLMs to generate Python code. This approach leverages the LLMs' strong Python coding capabilities and existing Python libraries for temporal calculations. The generated code serves as an intermediate rationale. By executing the code line-by-line, the model follows step-by-step reasoning similar to CoT \cite{wei2022chain}, leading the model to select the correct answer.

Due to the absence of temporal reasoning evaluation benchmarks specific to multi-session dialogues, we propose a method to construct a new evaluation benchmark focusing on two key temporal characteristics: \textit{relative time} and \textit{cross-session dependency}. By augmenting dialogues from LoCoMo \cite{maharana-etal-2024-evaluating}, we create multiple-choice temporal questions spanning three types of reasoning to evaluate the temporal reasoning capabilities of LLMs in this context.

We evaluate our framework based on three popular LLMs—GPT-4o, GPT-4o-mini, and GPT-3.5-Turbo—on our benchmark. The results show that our benchmark is challenging, revealing suboptimal performance for LLMs. In contrast, our framework demonstrates superior performance compared to baseline methods, such as CoT, highlighting the effectiveness of our approach in improving temporal reasoning in multi-session dialogues.

Our contributions are as follows: 
\begin{itemize} 
    \item We propose a new framework for temporal reasoning in multi-session dialogues, integrating time-aware memorization and neuro-symbolic temporal reasoning. 
    \item We propose a method to construct a temporal reasoning evaluation benchmark for multi-session dialogues by augmenting an existing dataset, explicitly covering the temporal characteristics of relative time and cross-session dependency. 
    \item Through extensive experiments, we empirically show that temporal reasoning in multi-session dialogues poses significant challenges for LLMs, even with strategies like CoT. However, our framework significantly improves LLMs' temporal reasoning in this context. 
\end{itemize}

\begin{table*}[ht]
\small
\centering
\scalebox{0.86}{
\begin{tabular}{lp{1.44cm}p{1.75cm}p{1.6cm}p{2.1cm}p{3.5cm}}
\toprule
\bf Dialogue Dataset & \bf Avg. Turns Per Conv. & \bf Avg. Sessions Per Conv. & \bf Avg. Tokens Per Conv. & \bf Time Interval & \bf Collection \\ \midrule
MSC \cite{xu-etal-2022-beyond} & 53.3 & 4 & 1,225.9 & few days & Crowdsourcing \\
Conversation Chronicles \cite{jang2023conversation} & 58.5 & 5 & 1,054.7 &few hours - years & LLM-generated \\
\bf LoCoMo \cite{maharana-etal-2024-evaluating} (Ours) &304.9 & 19.3 & 9,209.2 & few months & LLM-gen. + crowdsourcing \\
 \bottomrule
\end{tabular}}
\caption{Statistics of the chosen multi-session dialogue dataset, LoCoMo, compared to others.}
\label{tab:locomo_statistics}
\end{table*}

\section{Benchmark Construction}
In this section, we introduce the construction pipeline to build our temporal QA benchmark for evaluating LLM-agents' temporal reasoning in multi-session dialogues. As mentioned earlier, we mainly focus on the two temporal characteristics in multi-session dialogues: \textbf{relative time} and \textbf{cross-session dependency}.

\begin{table*}[ht]
\small
\centering
\scalebox{1.0}{
\begin{tabular}{lllll}
\toprule
\bf Question Type & \bf Count & \bf $\#$ of Options & \bf $\#$ of Events & \bf Event Type \\ \midrule
Temporal Anchoring & 264 & 5 & 1 & relative time \\
Temporal Precedence & 102 & 3 & 2 & cross session dependency (+ relative time) \\
Temporal Interval & 234 & 5 & 2 & cross session dependency (+ relative time) \\ \midrule
Total & 600 & -- & -- & -- \\
\phantom{Total} - Unanswerable & 112 & -- & -- & -- \\ \midrule
LoCoMo \cite{maharana-etal-2024-evaluating} & 321 & -- & -- & -- \\ \bottomrule
\end{tabular}}
\caption{Dataset statistics and details of the constructed benchmark.\textit{$\#$ of Options} refers to the number of options for each temporal question. \textit{$\#$ of Events} refers to the number of selected events to create each temporal question. \textit{Event Type} specifies the type of temporal events chosen for question creation, where \textit{(+ relative time)} indicates that relative time was an additional consideration in event selection.  Note that our benchmark not only contains more temporal QAs than LoCoMo, but also include unanswerable questions.}
\label{tab:statistics}
\end{table*}

\subsection{Benchmark Design}
We propose augmenting an existing dataset to create a benchmark for evaluating LLM-agents' temporal reasoning in multi-session dialogues. After a thorough examination, we selected LoCoMo \cite{maharana-etal-2024-evaluating}, which comprises dialogues averaging 600 turns and 16,000 tokens across up to 32 sessions. In comparison to existing multi-session dialogue datasets, LoCoMo features the longest dialogues and the most sessions (as shown in Table \ref{tab:locomo_statistics}), thus presenting a greater challenge.

As mentioned earlier, our benchmark focuses on two key temporal characteristics in multi-session dialogues: \textit{relative time} and \textit{cross-session dependency}. To achieve this, we follow previous benchmarks \cite{chen2dataset,xiong-etal-2024-large,wang-zhao-2024-tram} by creating temporal QA pairs based on temporal events in the dialogues. Specifically, we design each temporal QA based on either a single event or a pair of events:

$\bullet$ \textbf{Single Event}: We select events expressed with relative time and develop a temporal reasoning type called \textit{Temporal Anchoring}, which asks for the exact time of the event.

$\bullet$ \textbf{Two Events}: We choose pairs of relevant events from different sessions that exhibit cross-session dependency. We also consider relative time as an extra factor to increase the complexity of the questions. Two temporal reasoning types are applied: \textit{Temporal Precedence}, which asks which event occurred first, and \textit{Temporal Interval}, which asks for the duration between the two events.

\subsection{Construction Pipeline}
To construct our benchmark, we follow the design of our benchmark and utilize a systematic step-by-step approach with GPT-4o. 
The prompt for each step is shown in Appendix $\S$ \ref{sec:benchmark_prompt}.

\noindent \textbf{Step 1: Temporal Event Extraction}
We begin by prompting GPT-4o to extract all temporal events from each dialogue session. In addition, we instruct GPT-4o to annotate the relative time expressions for these events, facilitating the selection process for creating temporal QAs.

\noindent \textbf{Step 2: Temporal Event Linking}
Next, we link the extracted events containing cross-session dependency within the dialogue. We prompt GPT-4o with the extracted events and instruct it to group those related to the same or relevant entities across different sessions, particularly those reflecting changes in attributes over time. For example, the event “Debra Ryan told her mentor about her business idea” from an early session is linked to “Debra Ryan started her own business” from a later session.

\noindent \textbf{Step 3: Temporal QA Creation}
Once the temporal events are processed, we prompt GPT-4o to select those events that meet the criteria for different temporal reasoning types and generate multiple-choice temporal QAs. Additionally, we create unanswerable questions, as in prior QA benchmarks \cite{rajpurkar2018know}, to comprehensively assess models' temporal reasoning capabilities.

\noindent \textbf{Step 4: Quality Control}
We observe various noises in generated QAs, such as incorrect inferences about exact times. To ensure the benchmark’s quality, we follow recent temporal reasoning benchmarks for LLMs, such as TGQA \cite{xiong-etal-2024-large}, to perform quality control. We manually review each question to verify that it aligns with our design specifications and that the answers are correctly grounded in the dialogue. We also revise well-constructed questions with incorrect answers and remove any unreasonable ones. The final temporal QA benchmark covers time intervals from days to months and its statistics and details are presented in Table \ref{tab:statistics}. Particularly, our final benchmark not only contains more temporal QAs than LoCoMo, but also include unanswerable questions, which are not covered in LoCoMo. We also include examples of QAs for different temporal reasoning types in Appendix $\S$. \ref{sec:QAs}.

\section{Methodology}

\subsection{Preliminary: Memory-Augmented LLM-Agents}
To address the limit of LLMs struggling in retaining information from long input text, recent studies turn to equip LLM agents with memory to support long-turn conversations \cite{lu2023memochat,packer2023memgpt,zhong2024memorybank}. Therefore, we base our study on memory-augmented LLM-agents.

The general pipeline of memory-augmented LLM-agents comprises three stages: \textit{memorization}, \textit{retrieval}, and \textit{response}. In the \textit{memorization} stage, the model summarizes each dialogue session and stores these summaries as memory. During the \textit{retrieval} stage, the model retrieves the most relevant memory for the current dialogue session. This retrieved memory is then concatenated with the ongoing dialogue to generate the next \textit{response}.
Specifically, we build our framework based on MemoChat \cite{lu2023memochat}, which realizes this three-stage process through prompting and has demonstrated effectiveness in handling long-range dialogues.

\subsection{TReMu}
Building on the memory-augmented LLM-agent pipeline, we introduce our framework called \textbf{TReMu} as shown in Algorithm \ref{algo1}. The framework consists of two key components: \textit{time-aware memorization} and \textit{neuro-symbolic temporal reasoning}.

\begin{algorithm}
\small
\caption{TReMu} \label{algo1}
\begin{algorithmic}
\STATE \textbf{Initialize:} 
\STATE Time-aware Memorization Model LLM\textsubscript{mem}
\STATE Memory Retrieval Model LLM\textsubscript{retrieval}
\STATE Neuro-symbolic Reasoning Model LLM\textsubscript{code}
\STATE Symbolic Solver $\mathcal{P}$
\STATE Memorization pool $\mathcal{M} \leftarrow \emptyset $
\STATE \COMMENT{\textit{Time-aware Memorization}}
\FOR{each dialogue session $d_i$ in dialogue $\mathcal{D}$}
    \STATE \( m_i \leftarrow \text{LLM}_\text{mem}(d_i) \)
    \STATE \( \mathcal{M} \leftarrow f_{org} ( \mathcal{M}, m_i ) \)
\ENDFOR
\STATE \COMMENT{\textit{Neuro-symbolic Temporal Reasoning}}
\FOR{each temporal question $q$}
\STATE  \(m_{retrieved} \leftarrow \text{LLM}_\text{retrieval}(q, \mathcal{M})\)
\STATE \(c \leftarrow \text{LLM}_{code}(q, m_{retrieved})\)
\STATE \(o \leftarrow \mathcal{P}(c)\)
\STATE final answer \(a \leftarrow \text{LLM}(q, o)\)
\ENDFOR
\end{algorithmic}
\end{algorithm}

\subsubsection{Time-aware Memorization} 

Our time-aware memorization builds on timeline summarization \cite{steen-markert-2019-abstractive,rajaby-faghihi-etal-2022-crisiltlsum,sojitra2024timeline} and it consists of two steps: \textit{Temporal Memory Writing} and \textit{Memory Organization}. During Temporal Memory Writing (prompt in Appendix $\S$.\ref{sec:timeline_prompt}), we instruct LLM agents to generate memory pieces while also extracting and associating mentioned events with inferred dates. Unlike prior approaches that summarize entire sessions holistically, our method produces fine-grained memory pieces linked to specific inferred time markers. As shown in Tables \ref{tab:memochat} and \ref{tab:timeaware}, our memorization outputs memory pieces corresponding to events with inferred time steps that facilitates to mitigate temporal ambiguity. For example, the highlighted texts show that Michelle cooked a meal and later referenced cooking it at different dates. This enables finer temporal granularity, effectively distinguishing events based on inferred time intervals.

\begin{table}[h]
    \centering
    \scalebox{1.0}{
    \begin{tabular}{p{2.2cm} p{4.5cm}}
        \toprule
        \textbf{Topic} & \textbf{Summary} \\
        \midrule
        Catching Up & Daniel and Michelle catch up on new events in their lives including new jobs, hobbies, and activities. \\
        Hobbies and Daily Rituals & Daniel and Michelle discuss their hobbies and rituals like running, ballet, playing guitar, meditation, and cooking. \\
        Cooking and Celebrations & Both talk about their cooking experiences and celebrate Daniel's promotion. \\
        Books and Recommendations & Michelle and Daniel discuss books they've read and recommend some to each other. \\
        Personal Items with Sentimental Value & Michelle and Daniel talk about cameras and a vintage motorcycle with sentimental value. \\
        \bottomrule
    \end{tabular}}
    \caption{Output memory from MemoChat based on one dialogue session in LoCoMo.}
    \label{tab:memochat}
\end{table}

\begin{table}[h]
    \centering
    \scalebox{1.0}{
    \begin{tabular}{p{1.5cm} p{5.2cm}}
        \toprule
        \textbf{Time} & \textbf{Summary} \\
        \midrule
        01/28/2020 & Daniel and Michelle share updates on their lives...
        including Michelle starting her Masters in Psychology, Daniel starting a new job where he learns to code and problem-solve, and Michelle's hobby of ballet, meditation, and journaling...
        \textbf{Michelle mentions she made an Italian meal last Saturday} and Daniel made salsa for a taco night. Daniel also shared receiving a promotion ... 
        \\
        01/27/2020 & Daniel received a promotion and celebrated with a dinner at his favorite spot. \\
        01/25/2020 & \textbf{Michelle cooked a delicious Italian meal for her friends}, including pasta, garlic bread, and tiramisu. \\
        01/24/2020 & Daniel made a huge batch of salsa and hosted a taco night with friends. \\
        01/20/2020 & Michelle started her Masters in Psychology. \\
        \bottomrule
    \end{tabular}}
    \caption{Output memory from Time-aware Memorization based on the same dialogue session in LoCoMo.}
    \label{tab:timeaware}
\end{table}

Then, we perform Memory Organization on the output memory pieces to maintain long-term memory. We structure memory in a timeline format, grouping events that occur simultaneously and indexing them based on inferred timesteps. This approach enhances the distinction between an event’s occurrence and its mention, reducing temporal ambiguity and improving time-based retrieval. These enhancements mark a significant difference from traditional memorization approaches, supporting efficiency in temporal reasoning.

\subsubsection{Neuro-symbolic Temporal Reasoning}


Inspired by the recent progress in neuro-symbolic reasoning for LLMs \cite{han2022folio, pan2023logic}, we propose leveraging LLMs to translate temporal reasoning questions into Python code as intermediate rationales, which is executed as the reasoning process to derive answers (prompts shown in Appendix $\S$.\ref{sec:reasoning_prompt}). We tried different symbolic languages and finally chose Python because SOTA LLMs are better at generating Python code and there exist Python libraries that support temporal calculations, like \textit{datetime} and \textit{dateutil}. Particularly, \textit{dateutil} provides a function \textit{relativedelta} supporting relative time calculation, for example we can infer next Friday using \textit{TODAY + relativedelta(weekday=FR)}. Meanwhile, we provide implemented functions to be directly called, such as "weekRange(\textit{t})" returns the start date and the end date of the week that \textit{t} lies in.  Different from other works in temporal reasoning based on code execution\cite{li2023unlocking}, we enable our LLM agents with \textbf{function-calling} abilities, ensuring correctness and expanding the range of temporal reasoning tasks beyond simple precedence relations.

We provide demonstration via in-context learning to generate Python code with function calling, given then question and retrieved memory. After the generated code is executed, the output and code serve as \textbf{intermediate rationales}, and the LLM is prompted again to give the answer. Particularly, our reasoning approach offers an alternative form of CoT. While the original CoT \cite{wei2022chain} performs step-by-step reasoning in natural language, our neuro-symbolic approach conducts temporal reasoning by executing generated code line-by-line in a programming language. This neuro-symbolic method retains the core concept of CoT’s step-by-step reasoning while leveraging the precision and additional support provided by Python code and packages. However, prior works \cite{li2023unlocking} rely solely on solver outputs without providing intermediate justifications.

\begin{table*}[ht]
\small
\centering
\scalebox{1.0}{
\begin{tabular}{lccccccc}
\toprule
 \multicolumn{1}{c}{\multirow{2}{*}{\bf Methods}} & \multicolumn{4}{c}{\bf Accuracy} & \multicolumn{3}{c}{\bf Unanswerable Questions} \\ \cmidrule(r){2-5} \cmidrule(r){6-8} & TA & TP & TI & Overall & Precision & Recall & F1 \\ \midrule 
 SP & 18.18 & 58.82 & 30.34 & 29.83 & 46.88 & 13.39 & 20.84 \\
 CoT & 67.80 & 74.51 & 49.15 & 61.67 & 42.61 & 43.75 & 43.18 \\
 MemoChat & 35.23 & 43.14 & 25.21 & 32.67 & 24.30 & 77.68 & 37.02 \\
 Memochat + CoT & 51.14 & 49.02 & 26.50 & 41.67 & 24.80 & \bf 81.25 & 38.00 \\
 Timeline + CoT & 83.33 & 78.41 & 58.55 & 71.50 & 48.51 & 58.04 & 52.84 \\
 TReMu & \bf 84.47 & \bf 81.37 & \bf 68.38 & \bf 77.67 & \bf 55.48 & 76.79 & \bf 64.42\\
\bottomrule

\end{tabular}}
\caption{\label{gpt4o_result} Experimental results of various methods based on GPT-4o. We use TA to represent Temporal Anchoring, TP for Temporal Precedence and TI for Temporal Interval.}
\end{table*}

\begin{table*}[ht]
\small
\centering
\scalebox{1.0}{
\begin{tabular}{lccccccc}
\toprule
 \multicolumn{1}{c}{\multirow{2}{*}{\bf Methods}} & \multicolumn{4}{c}{\bf Accuracy} & \multicolumn{3}{c}{\bf Unanswerable Questions} \\ \cmidrule(r){2-5} \cmidrule(r){6-8} & TA & TP & TI & Overall & Precision & Recall & F1 \\ \midrule 
 SP & 20.08 & 50.00 & 29.91 & 29.00 & \bf 40.00 & 26.79 & 32.08 \\
 CoT & 46.59 & \bf 62.75 & 37.18 & 45.67 & 33.96 & 48.21 & 39.86 \\
 MemoChat & 21.21 & 39.22 & 23.50 & 25.17 & 21.11 & 74.11 & 32.88 \\
 Memochat + CoT & 24.62 & 45.10 & 24.36 & 28.00 & 21.11 & 75.00 & 32.94 \\
 Timeline + CoT & 55.68 & 59.80 & 38.46 & 49.67 & 30.73 & 59.82 & 40.60 \\
 TReMu & \bf 64.02 & 46.08 & \bf 38.89 & \bf 51.17 & 29.21 & \bf92.86 & \bf 44.44\\
\bottomrule

\end{tabular}}
\caption{\label{gpt4o_mini_result} Experimental results of various methods based on GPT-4o-mini.}
\vspace{-0.5cm}
\end{table*}

\begin{table*}[ht]
\small
\centering
\scalebox{1.0}{
\begin{tabular}{lccccccc}
\toprule
 \multicolumn{1}{c}{\multirow{2}{*}{\bf Methods}} & \multicolumn{4}{c}{\bf Accuracy} & \multicolumn{3}{c}{\bf Unanswerable Questions} \\ \cmidrule(r){2-5} \cmidrule(r){6-8} & TA & TP & TI & Overall & Precision & Recall & F1 \\ \midrule 
 SP & 21.59 & 31.37 & 23.08 & 23.83 & 22.91 & 46.43 & 30.68 \\
 CoT & 23.86 & 38.24 & 22.65 & 25.83 & 20.97 & 50.00 & 29.56 \\
 MemoChat & 17.42 & 45.10 & 23.50 & 24.50 & 21.93 & 66.96 & 33.04 \\
 Memochat + CoT & 20.45 & \bf 53.92 & \bf 26.50 & 28.50 & 21.79 & 50.00 & 30.36 \\
 Timeline + CoT & 32.58 & 44.12 & 22.65 & 30.67 & 22.57 & 51.79 & 31.44 \\
 TReMu & \bf 42.42 & 37.25 & 22.22 & \bf 33.67 & \bf 23.33 & \bf 75 & \bf 35.60\\
\bottomrule

\end{tabular}}
\caption{\label{gpt35_result} Experimental results of various methods based on GPT-3.5-Turbo.}
\end{table*}

\section{Experiments}

\subsection{Experimental Setup} 
\noindent \textbf{Models.}
We build our framework using various \textit{black-box} LLMs: GPT-4o\footnote{Specifically, \textit{gpt-4o-2024-05-13}.}, GPT-4o-mini\footnote{Specifically, \textit{gpt-4o-mini-2024-07-18}.}, and GPT-3.5-Turbo\footnote{Specifically, \textit{gpt-3.5-turbo-0125}.}. Particularly, for GPT-3.5-Turbo, many of LoCoMo dialogues are longer than its input length, we then follow LoCoMo \cite{maharana-etal-2024-evaluating} which earlier dialogues are omitted. 

\label{sec:models}
Particularly, we have also tried different \textit{open-source} LLMs but most of them cannot handle the long dialogue inputs from LoCoMo, for example only about 10\% of dialogues can be fed into Llama-3-70B. And even for those shorter dialogues that can be fed into Llama-3-70B, we notice that the model gets lost and fails to follow instructions, even failing to generate in the desired format. Therefore, we leave the adaptation to LLMs as future work.

\noindent\textbf{Baselines.}
Since in our setting of multi-session dialogues where the conversations exceed the input limits of LLMs, we consider the memory mechanism as a critical component of baselines in order to feed complete dialogue information. Therefore, we include the following baselines for comparison:

\begin{itemize} 
    \item \textbf{Standard Prompting (SP)}: The entire dialogue is provided along with each temporal question, with additional instructions for selecting the correct answer.

    \item \textbf{Chain-of-Thought (CoT)} \cite{wei2022chain}: Similar to SP, but with additional instructions for LLMs to solve questions step-by-step.

    \item \textbf{MemoChat} \cite{lu2023memochat}: Given that multi-turn dialogues can exceed the model's input length, and since our approach builds on memory-augmented LLM-agents, MemoChat serves as a baseline where we modify the response stage to answer temporal questions.
\end{itemize}

To better understand the effectiveness of each component in our framework, we evaluate the following variants as baselines for the ablation study:
\begin{itemize}
    \item \textbf{MemoChat + CoT}: This baseline applies CoT in the response stage to answer temporal questions step-by-step using the retrieved memory.

    \item \textbf{Timeline + CoT}: Based on the framework of memory-augmented LLM-agents, we modify the original memorization with our proposed timeline summarization and combine it with CoT as a baseline.
\end{itemize}

Comparing MemoChat + CoT and Timeline + CoT allows us to assess the impact of replacing the standard memory mechanism in LLM agents with our time-aware memorization. Additionally, comparing Timeline + CoT with TReMu highlights the effect of replacing CoT with our neuro-symbolic reasoning approach.

\noindent\textbf{Evaluation Metrics.}
We primarily use \textbf{accuracy} to assess the overall performance of temporal reasoning. In addition, for unanswerable questions, we calculate \textbf{precision}, \textbf{recall}, and the \textbf{F1} score to specifically measure performance on this subset of questions. Specifically, precision is computed as the accuracy of questions the model predicts as "unanswerable," while recall is determined by the accuracy of questions where the ground truth answer is "unanswerable."

\subsection{Experimental Results} 
The results are shown in Tables \ref{gpt4o_result}, \ref{gpt4o_mini_result}, and \ref{gpt35_result} for GPT-4o, GPT-4o-mini, and GPT-3.5-Turbo, respectively. On the recent TRAM benchmark \cite{wang-zhao-2024-tram}, existing LLMs demonstrate strong performance with direct prompting. For instance, GPT-4 achieves an accuracy of 82 using CoT, while GPT-3.5 attains 71.40. In contrast, our benchmark is significantly more challenging. GPT-4o achieves only 61.67 with CoT and 29.83 with SP, whereas GPT-3.5 performs even worse, scoring 25.83 with CoT and 23.83 with SP. These performance gaps likely stem from the complexity of multi-session dialogues and their temporal dependencies, which are not explicitly addressed in previous benchmarks.

Our framework outperforms all baseline methods across all three LLMs in terms of both accuracy and F1 scores, with a notable increase in accuracy from 29.83 with SP to 77.67 with our framework using GPT-4o. This demonstrates the effectiveness of our approach in enhancing temporal reasoning for multi-session dialogues. However, incorporating a memory mechanism performs worse than CoT for GPT-4o and GPT-4o-mini. This may be because these models have sufficient input lengths to process LoCoMo dialogues, enabling them to identify relevant temporal information without additional memory augmentation. In contrast, for GPT-3.5, which has a shorter input limit, the memory mechanism generally improves performance by allowing the model to retrieve information from memory rather than truncated dialogues. 


Furthermore, we find that incorporating CoT generally improves performance, aligning with previous findings \cite{wang-zhao-2024-tram,xiong-etal-2024-large}. In particular, CoT encourages models to search for relevant information within dialogues. However, due to the dialogues’ length, the models sometimes generate responses like "I cannot find the mention of ...," which hinders their temporal reasoning capabilities. This further underscores the necessity of a memory mechanism to support long dialogue settings.

\begin{figure}  
\centering
\includegraphics[width=0.85\linewidth]{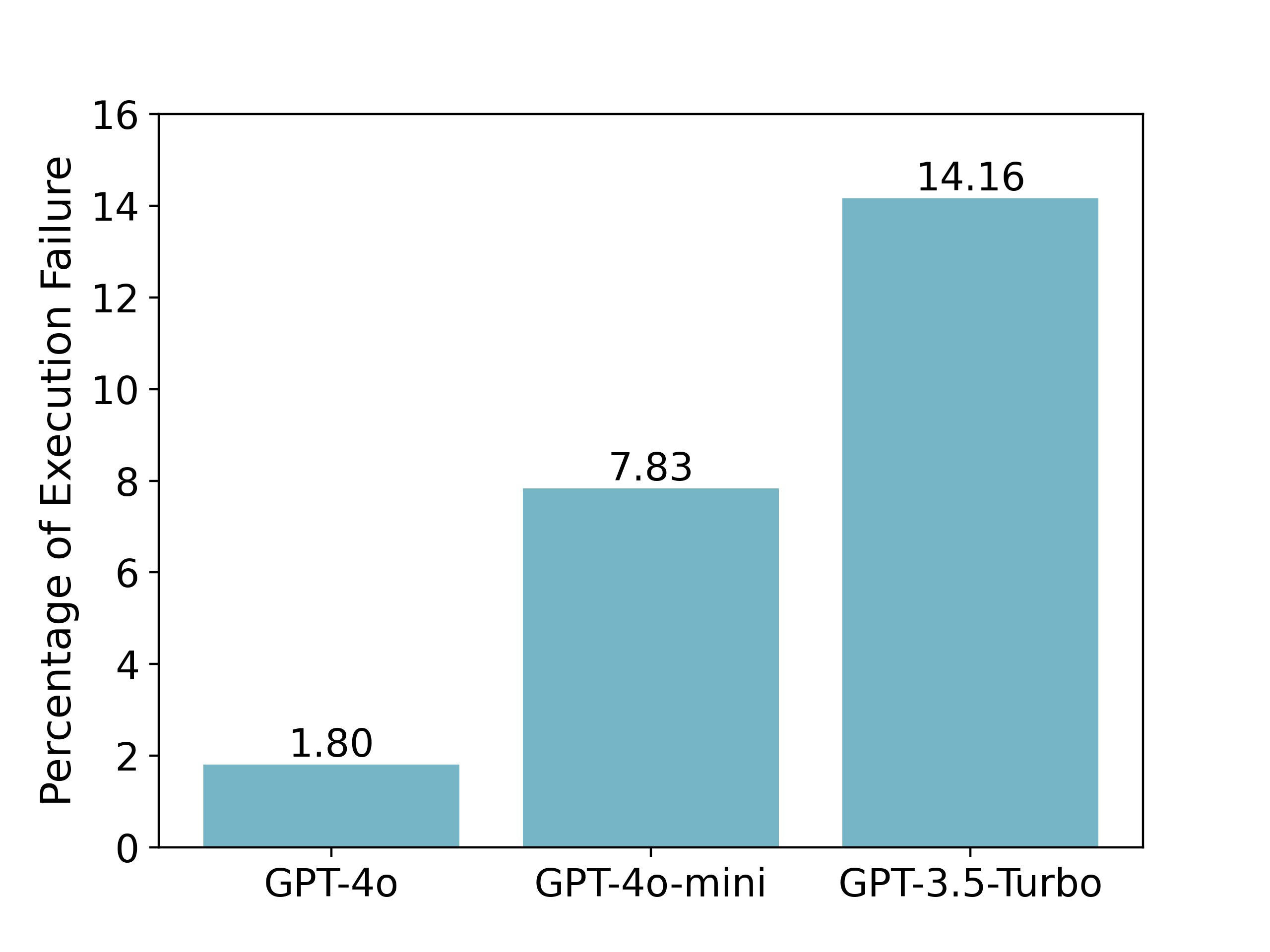}
    \caption{The percentage of execution failures.} 
    \label{fig:failure}
\end{figure}

\begin{figure*}[t]
    \centering
    \includegraphics[width=\linewidth]{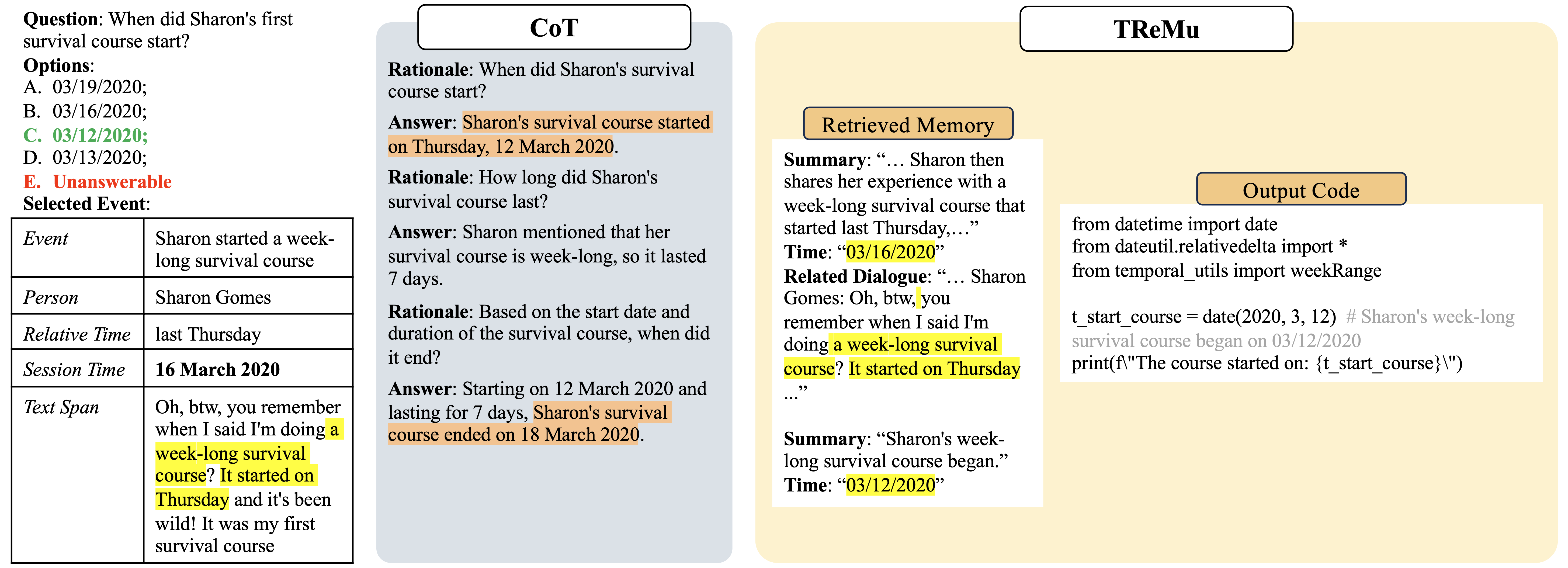}
    \caption{The first case study comparing CoT and our proposed framework, where CoT results in the wrong answer "E" but our approach selects the correct option "C". We highlight the key information in colors.}
    \label{fig:case_1}
\end{figure*}

\begin{figure*}[t]
    \centering
    \includegraphics[width=\linewidth]{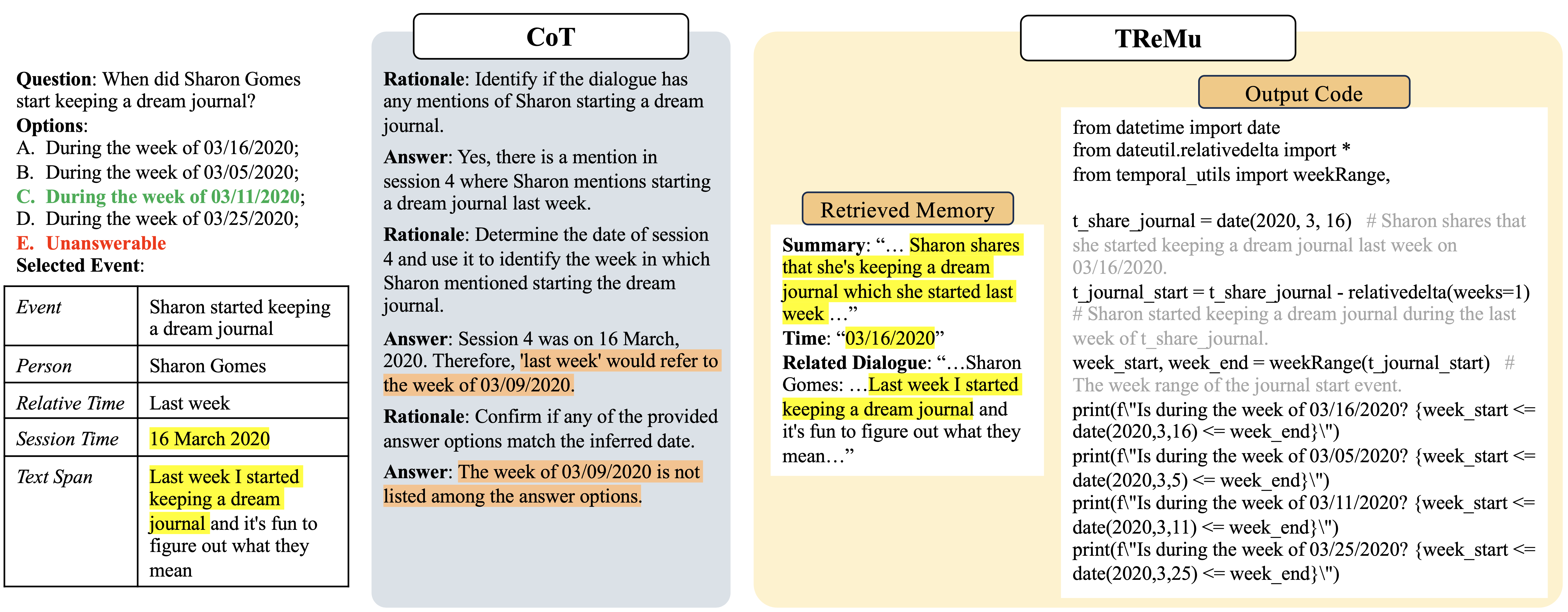}
    \caption{The second case study comparing CoT and our proposed framework, where CoT results in the wrong answer "E" but our approach selects the correct option "C". We highlight the key information in colors.}
    \label{fig:case_2}
\end{figure*}

\subsection{Ablation Study}
From Tables \ref{gpt4o_result}, \ref{gpt4o_mini_result}, and \ref{gpt35_result}, the comparison between MemoChat + CoT and Timeline + CoT highlights the importance of memory representation. Time-aware memorization improves accuracy by instructing models to infer temporal information—particularly relative time—during memorization and mitigate temporal ambiguity. Furthermore, replacing CoT with symbolic reasoning, as seen in the comparison between Timeline + CoT and TReMu, leads to additional performance gains. This improvement stems from the models’ ability to generate Python code while retaining the benefits of step-by-step reasoning. This aligns with recent research integrating LLMs with symbolic reasoners for various reasoning tasks \cite{olausson2023linc,pan2023logic}.

\subsection{Execution Failure Study}
We also measure the percentage of generated code that fails to execute and, during inference, we regenerate the code when such errors occur. The results, shown in Figure \ref{fig:failure}, indicate that the percentages of execution failure are generally low across all three LLMs, demonstrating the reliability of our Python-based symbolic reasoning approach. As expected, GPT-4o exhibits the lowest rate of execution failure, while GPT-3.5-Turbo has the highest, corresponding to the overall performance differences we demonstrate above in temporal reasoning among these models. This likely reflects the inherent performance gap between the LLMs.

\subsection{Case Study}
In this section, we demonstrate how the two key components of our framework—\textit{time-aware memorization} and \textit{neuro-symbolic temporal reasoning}—work in real cases. We compare the outputs of CoT and our framework based on GPT-4o.

In Figure \ref{fig:case_1}, with CoT, even though GPT-4o successfully identifies that the key temporal information is "Sharon's survival course started on 12 March 2020," but it gets confused with "week-long course" and infers the end date of the course, incorrectly selecting "Unanswerable." In contrast, with our framework’s time-aware memorization, the model retrieves the event from memory along with its properly inferred time. During the reasoning stage, the model utilizes this memory to distinguish between when the speaker, Sharon, mentioned the event (03/16/2020) and when the event occurred (03/12/2020). Then the model defines the corresponding variable in the generated code, i.e., \textit{t$\_$start$\_$course}, to precisely capture the time.

Figure \ref{fig:case_2} illustrates another mistake made via CoT. The model correctly infers that the "last week" corresponds to the session time of 16 March 2020 but fails to match the week range with the correct answer—the week of 03/09/2020 is the week of 03/11/2020 but the model does not realize this. As for our framework, the model leverages the Python \textit{dateutil} package's \textit{relativedelta} function, alongside our custom \textit{weekRange} function, to accurately infer the last week's range. This neuro-symbolic reasoning not only facilitates the model to reason step-by-step but also enhances it by incorporating external temporal functions to support more accurate temporal reasoning.
\section{Related Work}

\paragraph{Temporal Reasoning for LLMs.} 
Recent advancements in large language models (LLMs) have brought significant improvements in reasoning capabilities \cite{huang2023towards}, leading to growing interest in temporal reasoning \cite{chu-etal-2024-timebench,qiu2024large}. Existing approaches primarily address this challenge through time-aware language modeling. For example, \citeauthor{kanashiro-pereira-2022-attention, tan2023towards} propose fine-tuning strategies to enhance temporal reasoning, while \citeauthor{zhou2021temporal,yang-etal-2023-upon} introduce auxiliary objectives to incorporate external temporal knowledge. However, studies such as \citeauthor{chu-etal-2024-timebench,qiu2024large} show that state-of-the-art LLMs still exhibit suboptimal performance in temporal reasoning with prompting techniques. Our framework differs from these works by utilizing memory-augmented LLM agents, enhancing memorization through timeline summarization, and integrating neuro-symbolic reasoning as an intermediate step for answering temporal questions.

\paragraph{Multi-session Dialogues.} 
Except from the massive research efforts on common dialogue generation \cite{ge2023should}, several studies have developed multi-session dialogues benchmarks. \citeauthor{xu-etal-2022-beyond} introduced MSC, the first multi-session dataset incorporating time intervals between sessions. Similarly, \citeauthor{bae-etal-2022-keep} proposed a dynamic memory management method to maintain up-to-date user information and introduced a Korean multi-session dialogue dataset. \citeauthor{jang2023conversation} created the CONVERSATION CHRONICLES dataset, designed for long-term conversations that integrate time intervals and detailed speaker relationships. More recently, \citeauthor{maharana-etal-2024-evaluating} introduced LoCoMo, a dataset featuring long-term and multi-modal dialogues. While our work is situated within this context, it specifically targets temporal reasoning, addressing the temporal characteristics of relative time and cross-session dependency, which have not been explicitly explored in prior research.
\section{Conclusion}
In this paper, we address the critical challenge of temporal reasoning in multi-session dialogues for LLM-agents. We present a method to construct a temporal reasoning evaluation benchmark by augmenting dialogues from LoCoMo and covering the temporal characteristics of relative time and cross-session dependency. Furthermore, we introduce a novel framework which combines time-aware memorization through timeline summarization with neuro-symbolic temporal reasoning by translating temporal questions into executable Python code. 
Through extensive evaluations, we demonstrate that our benchmark presents significant challenges, and our framework substantially outperforms baseline methods in improving temporal reasoning for multi-session dialogues.

\section{Limitations}
Our work has several limitations:

\noindent\textbf{Assessment in QA settings.} Our evaluation follows the standard practice of recent temporal reasoning benchmarks like \cite{zhou2019going, wang-zhao-2024-tram}, using a multiple-choice format for reliable assessment. A more ideal setting would be evaluating in generative dialogue settings. However, generative QA evaluations pose significant challenges due to variations in temporal expressions (e.g., "January 1st" vs. "01/01"), making exact match and F1 token score unreliable. Thus, we prioritize benchmark reliability and accuracy over a more ambitious generative QA setting. We leave the extension of the current benchmark to generative dialogue settings as future work

\noindent\textbf{Open-sourced Models.} Our current experiments mainly focus on close-sourced models. However, as we have pointed out in our Experiment section $\S$ \ref{sec:models}, we found most open-source LLMs cannot handle the long dialogue inputs from LoCoMo, for example only about 10\% of dialogues can be fed into Llama-3-70B. And even for those shorter dialogues that can be fed into Llama-3-70B, we notice that the model gets lost and fails to follow instructions, even failing to generate in the desired format. We therefore consider the adaptation of our framework to open-sourced models as future work. Note that we will release our code and dataset for reproducibility.


\bibliography{acl_latex}

\appendix



\section{Prompts for Benchmark Construction}
\label{sec:benchmark_prompt}
\begin{figure*}
    \centering
    \includegraphics[width=\linewidth]{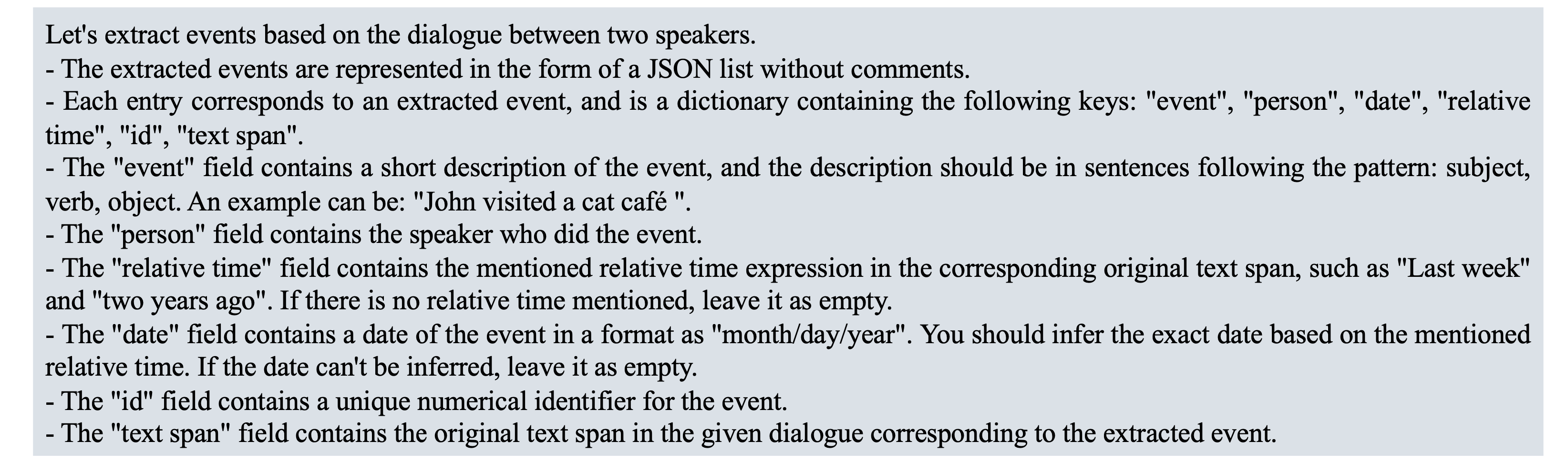}
    \caption{Prompt for Temporal Event Extraction.}
    \label{fig:event_extraction}
\vspace{-5mm}
\end{figure*}

\begin{figure*}
    \centering
    \includegraphics[width=\linewidth]{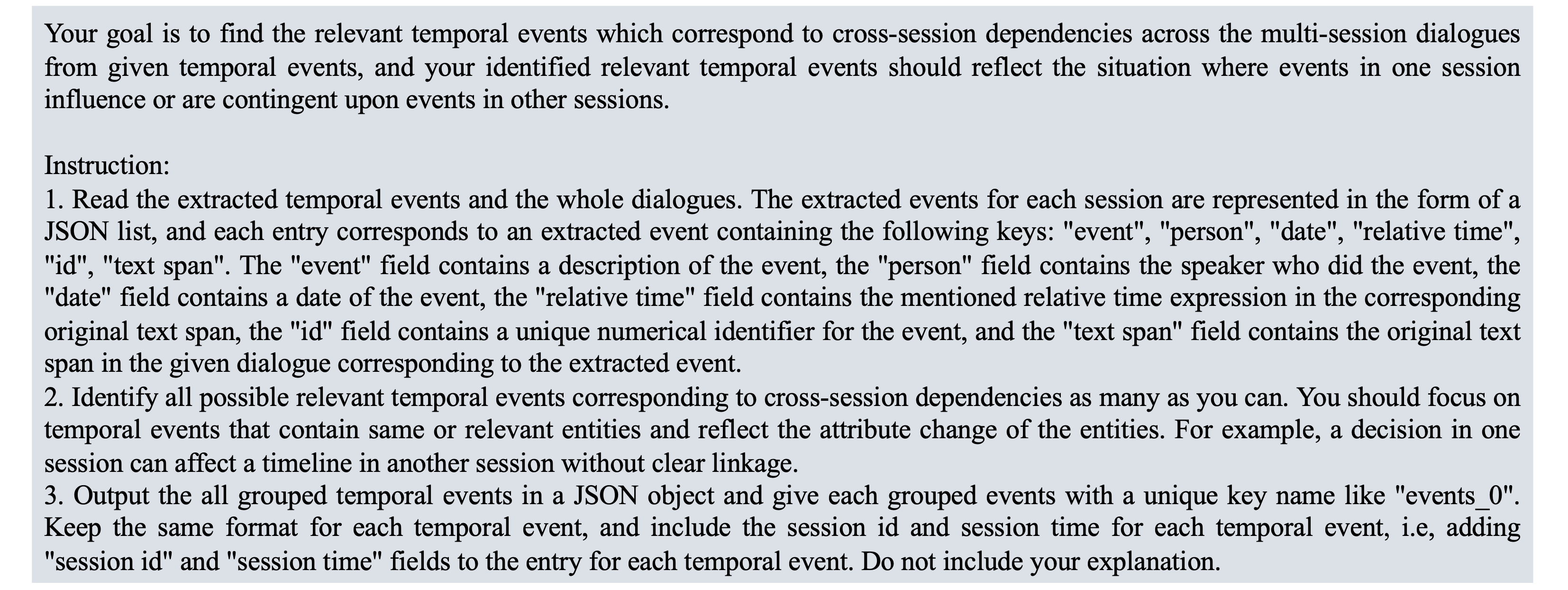}
    \caption{Prompt for Temporal Event Linking.}
    \label{fig:event_connection}
\vspace{-5mm}
\end{figure*}

\begin{figure*}
    \centering
    \includegraphics[width=\linewidth]{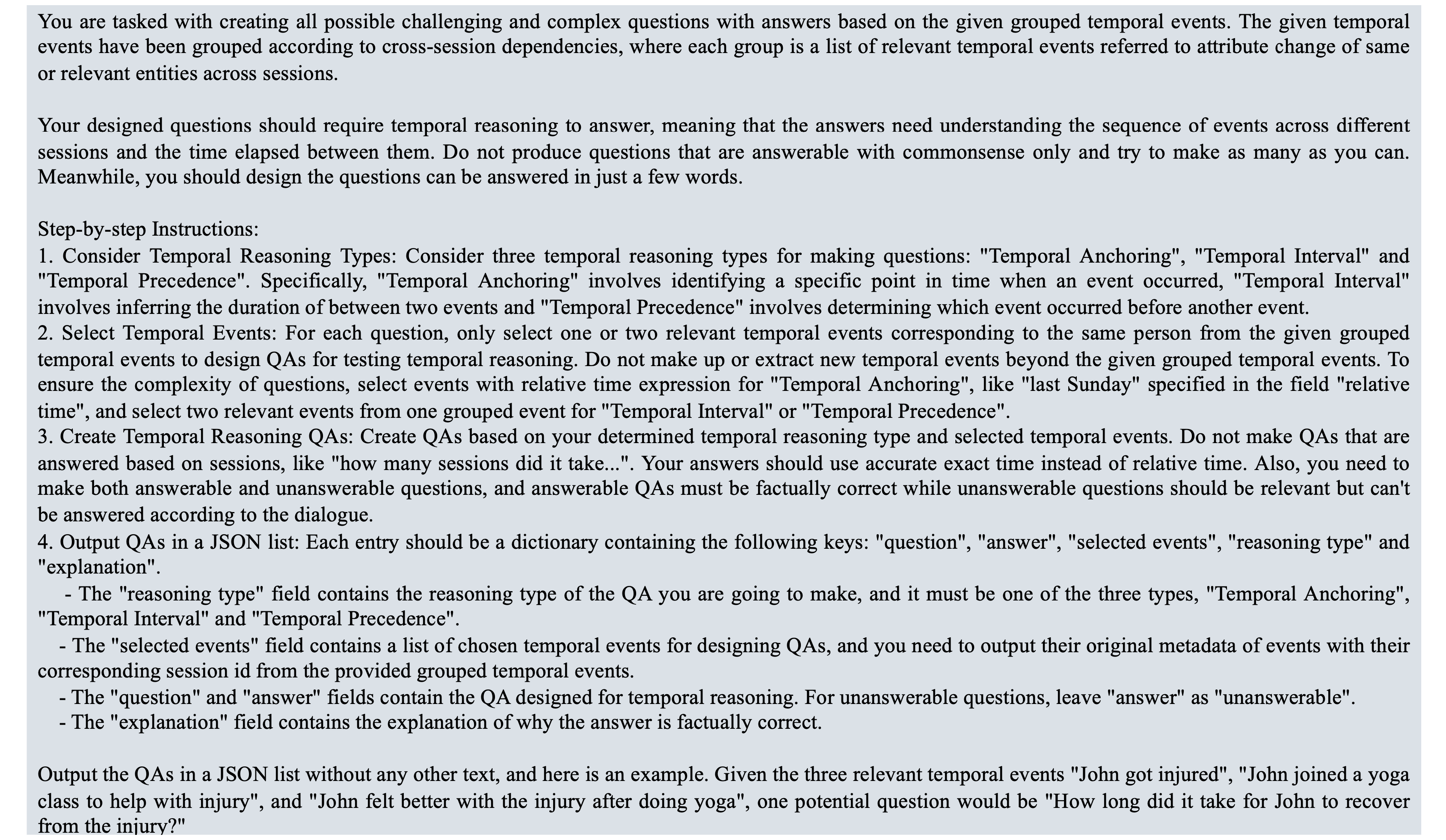}
    \caption{Prompt for Temporal QA Creation.}
    \label{fig:qa_creation}
\vspace{-5mm}
\end{figure*}

We use GPT-4o to construct a temporal reasoning benchmark for multi-session dialogues. The first step is the temporal event extraction using the prompt shown in Figure \ref{fig:event_extraction}. Then the prompt for the second step, temporal event linking, is shown in Figure \ref{fig:event_connection}. With the grouped temporal events, we use the prompt in Figure \ref{fig:qa_creation} to create temporal QAs.

\section{Examples of Temporal QAs in Constructed Benchmark}
\label{sec:QAs}

\begin{figure*}
    \centering
    \includegraphics[width=0.7\linewidth]{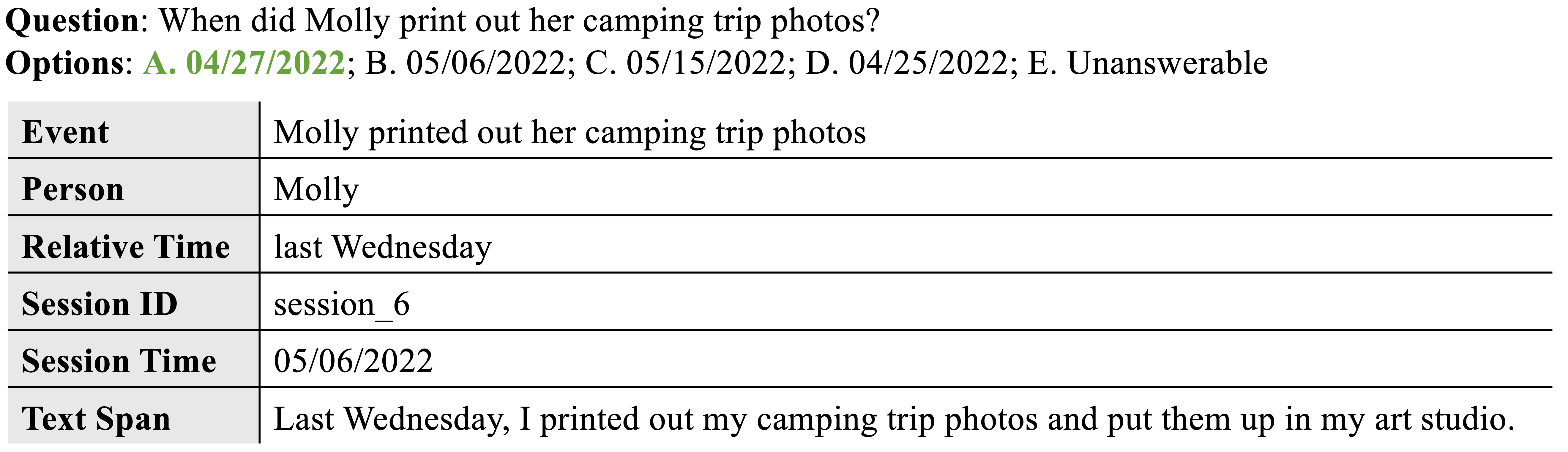}
    \caption{An example temporal QA for Temporal Anchoring}
    \label{fig:anchoring}
\end{figure*}

\begin{figure*}
    \centering
    \includegraphics[width=0.7\linewidth]{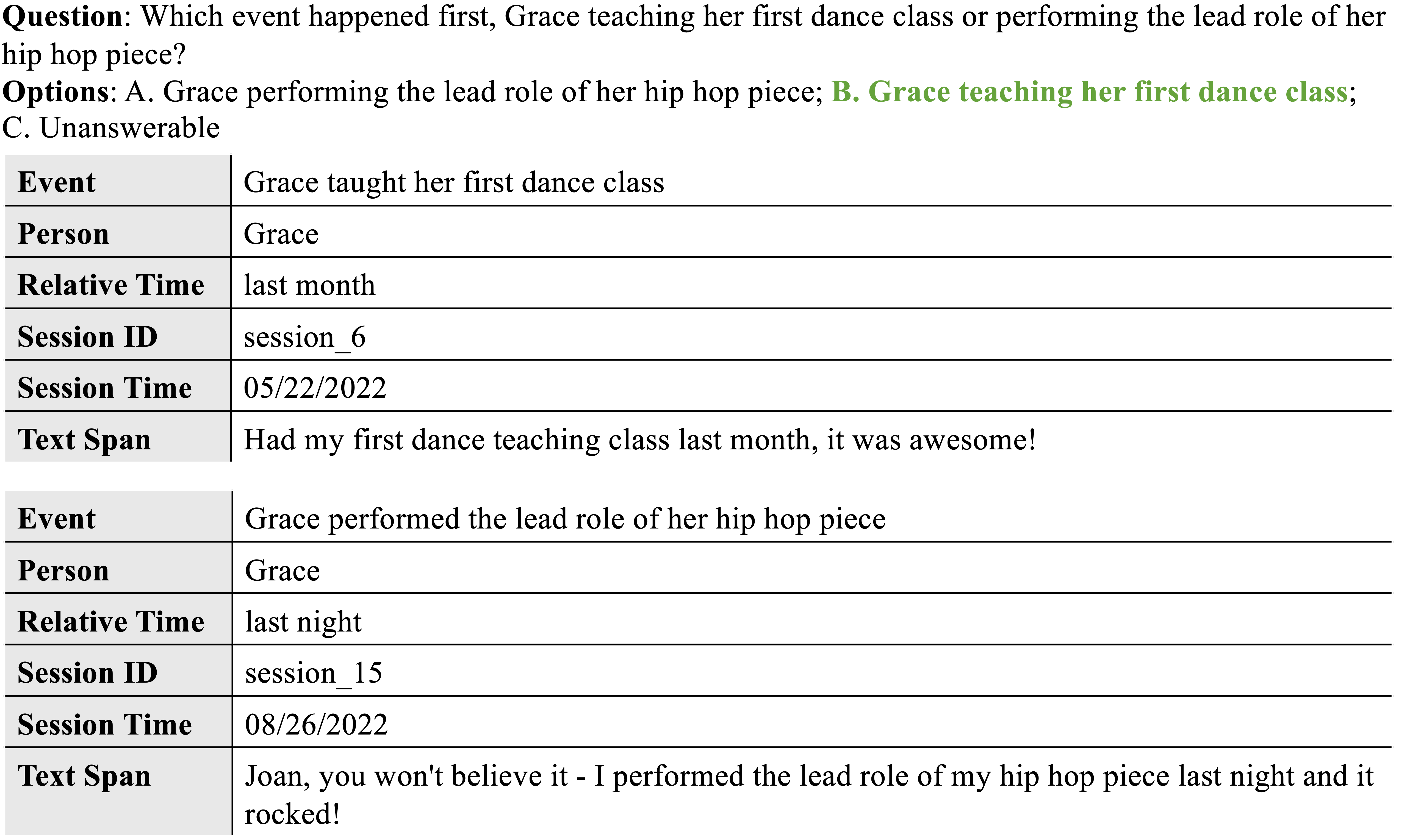}
    \caption{An example temporal QA for Temporal Precedence}
    \label{fig:precedence}
\end{figure*}

\begin{figure*}
    \centering
    \includegraphics[width=0.7\linewidth]{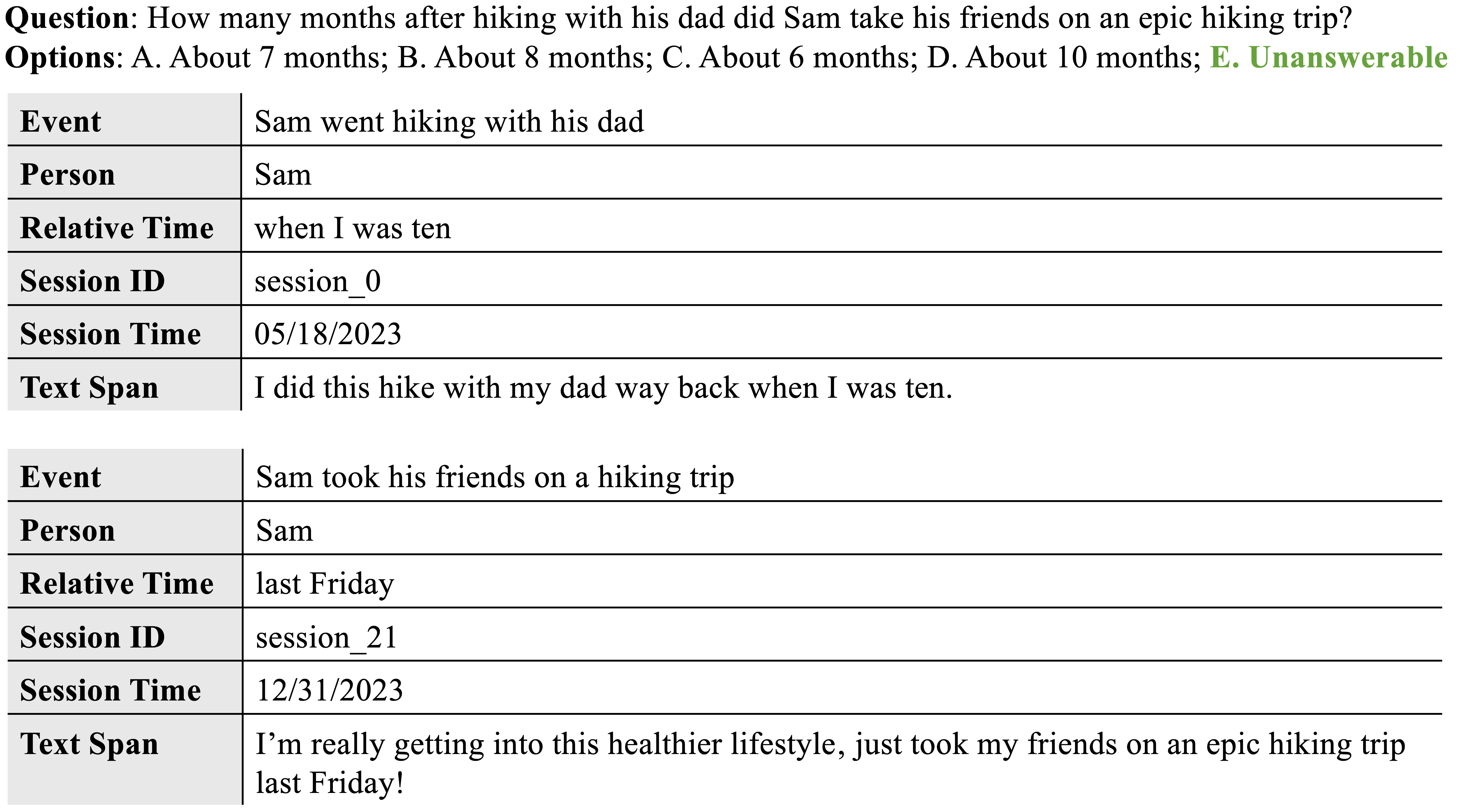}
    \caption{An example temporal QA for Temporal Interval}
    \label{fig:interval}
\end{figure*}

We show examples of final temporal QAs for different temporal reasoning types in Figure \ref{fig:anchoring}, \ref{fig:precedence} and \ref{fig:interval}. In each example, we highlight the ground truth answer as green and show the corresponding selected temporal events for constructing the question below the question.

\section{Prompt for Time-aware Memorization}
\label{sec:timeline_prompt}
\begin{figure*}
    \centering
    \includegraphics[width=\linewidth]{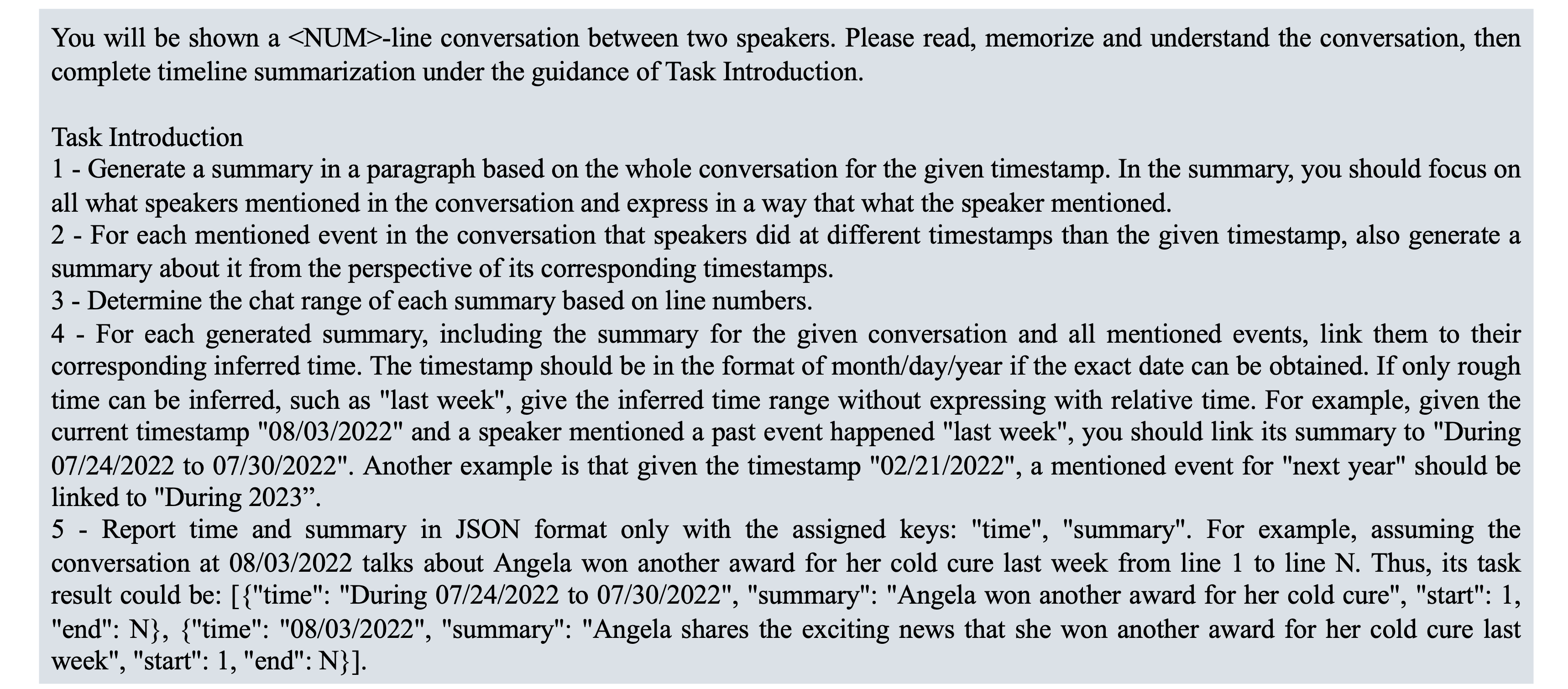}
    \caption{Prompt for Time-aware Memorization.}
    \label{fig:timeline_prompt}
\vspace{-5mm}
\end{figure*}

We show the designed prompt for time-aware memorization in Figure \ref{fig:timeline_prompt}.

\section{Prompt for Neuro-symbolic Temporal Reasoning}
\label{sec:reasoning_prompt}
\begin{figure*}
    \centering
    \includegraphics[width=\linewidth]{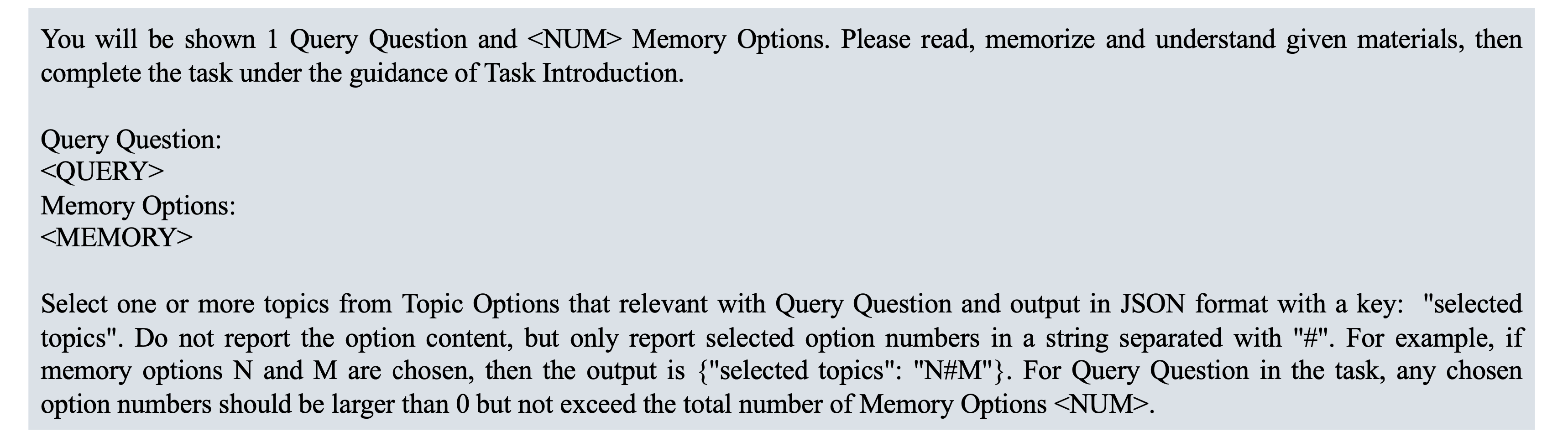}
    \caption{Prompt for memory retrieval.}
    \label{fig:retrieval_prompt}
\vspace{-5mm}
\end{figure*}

\begin{figure*}
    \centering
    \includegraphics[width=\linewidth]{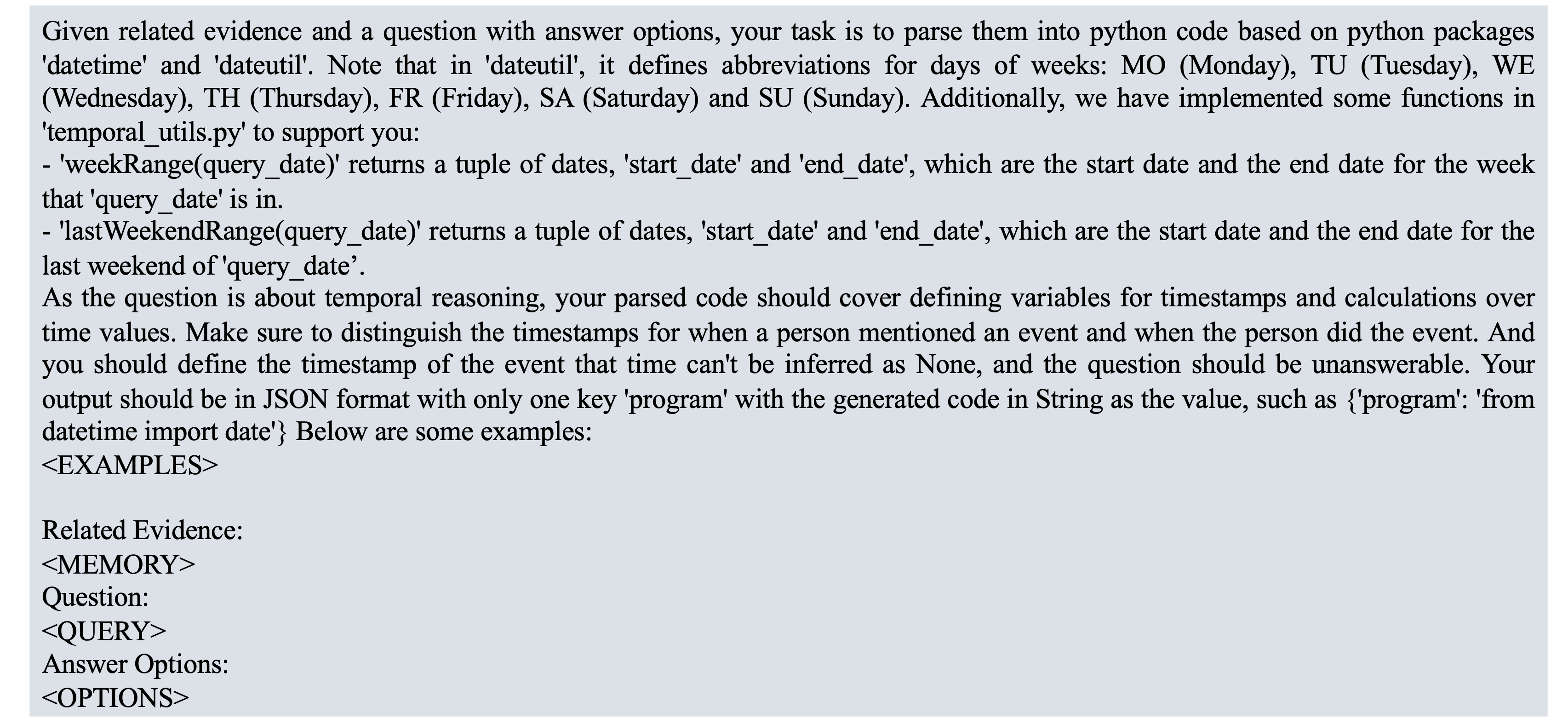}
    \caption{Prompt for generating Python code for temporal reasoning.}
    \label{fig:code_prompt}
\vspace{-5mm}
\end{figure*}

\begin{figure*}
    \centering
    \includegraphics[width=\linewidth]{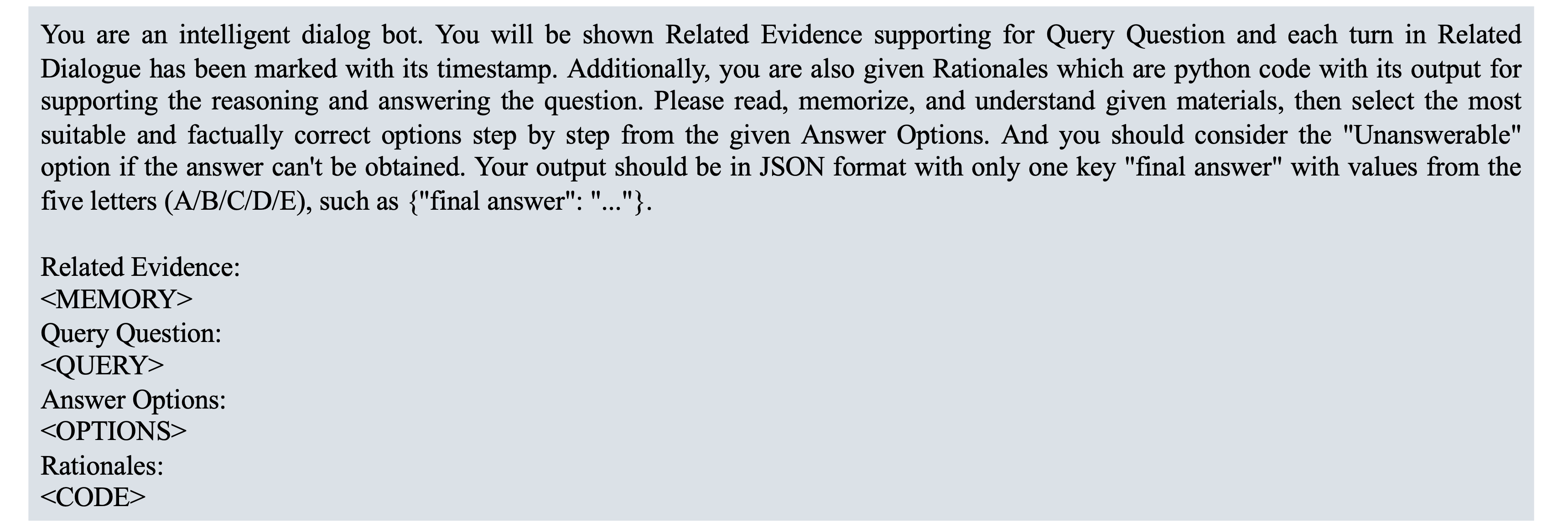}
    \caption{Prompt for temporal question answering.}
    \label{fig:qa_prompt}
\vspace{-5mm}
\end{figure*}

We show the designed prompts for neuro-symbolic temporal reasoning here. Specifically, we first perform memory retrieval with the prompt in Figure \ref{fig:retrieval_prompt}. Then we prompt to generate Python code via in-context learning as in Figure \ref{fig:code_prompt}. With the generated code and its execution result, we finally prompt the LLM to select the answer, and the prompt is shown in Figure \ref{fig:qa_prompt}.

\end{document}